\documentclass[10pt,twocolumn,letterpaper]{article}
\pdfoutput=1

\usepackage{cvpr}
\usepackage{times}
\usepackage{epsfig}
\usepackage{graphicx}
\usepackage{amsmath}
\usepackage{amssymb}

\usepackage{url}                                % simple URL typesetting
\usepackage{booktabs}                           % professional-quality tables
\usepackage{amsfonts}                           % blackboard math symbols
\usepackage{nicefrac}                           % compact symbols for 1/2, etc.
\usepackage{microtype}                          % microtypography
\usepackage[font=small,labelfont=bf]{caption}   % caption
\usepackage{bbm}                                % indicator
\usepackage{wrapfig}                            % wrap figure
\usepackage{color}                              % color text
\renewcommand{\*}[1]{\textbf{#1}}               % bold
\renewcommand{\u}{$\uparrow$}
\renewcommand{\d}{$\downarrow$}

\newcommand{\cut}[1]{}

\cvprfinalcopy % *** Uncomment this line for the final submission

\def\cvprPaperID{3051} % *** Enter the CVPR Paper ID here

\ifcvprfinal\fi
\begin{document}
\title{End-to-End Instance Segmentation with Recurrent Attention}

\author{
    Mengye Ren${}^1$, Richard S. Zemel${}^{1,2}$ \\
    University of Toronto${}^1$, Canadian Institute for Advanced Research${}^2$ \\
    {\tt \small \{mren,zemel\}@cs.toronto.edu} \\
}

\maketitle
\begin{abstract}
\noindent While convolutional neural networks have gained impressive success
recently in solving structured prediction problems such as semantic
segmentation, it remains a challenge to differentiate individual object
instances in the scene. Instance segmentation is very important in a variety of
applications, such as autonomous driving, image captioning, and visual question
answering. Techniques that combine large graphical models with low-level vision
have been proposed to address this problem; however,  we propose an end-to-end
recurrent neural network (RNN) architecture with an attention mechanism to
model a human-like counting process, and produce detailed instance
segmentations. The network is jointly trained to sequentially produce regions
of interest as well as a dominant object segmentation within each region. The
proposed model achieves competitive results on the CVPPP 
\cite{minervini14cvppp}, KITTI \cite{geiger12kitti}, and Cityscapes
\cite{cordts16cityscapes} datasets.
\end{abstract}

\input{figs/cvppp_total_fig.tex.part}
\section{Introduction}

Instance segmentation is a fundamental computer vision problem, which aims to
assign pixel-level instance labeling to a given image. While the standard
semantic segmentation problem entails assigning class labels to each pixel in an
image, it says nothing about the number of instances of each class in the image.
Instance segmentation is considerably more difficult than semantic segmentation
because it necessitates distinguishing nearby and occluded object instances.
Segmenting at the instance level is useful for many tasks, such as highlighting
the outline of objects for improved recognition and allowing robots to delineate
and grasp individual objects; it plays a key role in autonomous driving as well.
Obtaining instance level pixel labels is also an important step towards general
machine understanding of images.

Instance segmentation has been rapidly gaining in popularity, with a spurt of
research papers in the past two years, and a new benchmark competition, based on
the Cityscapes dataset \cite{cordts16cityscapes}.

A sensible approach to instance segmentation is to formulate it as a structured
output problem. A key challenge here is the dimensionality of the structured
output, which can be on the order of the number of pixels times the number of
objects. Standard fully convolutional networks (FCN) \cite{long15fcn} will have
trouble directly outputting all instance labels in a single shot. Recent work on
instance segmentation \cite{silberman14insseg, zhang15insseg,zhang16insseg}
proposes complex graphical models, which results in a complex and time-consuming
pipeline. Furthermore, these models cannot be trained in an end-to-end fashion.

One of the main challenges in instance segmentation, as in many other computer
vision tasks such as object detection, is occlusion.  For a bottom-up approach
to handle occlusion, it must sometimes merge two regions that are not connected,
which becomes very challenging at a local scale.  Many approaches to handle
occlusion utilize a form of non-maximal suppression (NMS), which is typically
difficult to tune.  In cluttered scenes, NMS may suppress the detection results
for a heavily occluded object because it has too much overlap with foreground
objects.  One motivation of this work is to introduce an iterative procedure to
perform dynamic NMS, reasoning about occlusion in a top-down manner.

A related problem of interest entails counting the instances of an object class
in an image. On its own this problem is also of practical value. For instance,
counting provides useful population estimates in medical imaging and aerial
imaging. General object counting is fundamental to image understanding, and our
basic arithmetic intelligence. Studies in applications such as image question
answering \cite{antol15vqa, ren15vqa} reveal that counting, especially on
everyday objects, is a very challenging task on its own
\cite{chattopadhyay16count}. Counting has been formulated in a task-specific
setting, either by detection followed by regression, or by learning
discriminatively with a counting error metric \cite{lempitsky10count}.

To tackle these challenges, we propose a new model based on a recurrent neural
network (RNN) that utilizes visual attention, to perform instance segmentation.
We consider the problem of counting jointly with instance segmentation. Our
system addresses the dimensionality issue by using a temporal chain that outputs
a single instance at a time. It also performs dynamic NMS, using an object that
is already segmented to aid in the discovery of an occluded object later in the
sequence. Using an RNN to segment one instance at a time is also inspired by
human-like iterative and attentive counting processes. For real-world cluttered
scenes, iterative counting with attention will likely perform better than a
regression model that operates on the global image level. Incorporating joint
training on counting and segmentation allows the system to automatically
determine a stopping criterion in the recurrent formulation we formulate here.

\section{Recurrent attention model} 

\begin{figure*}
\begin{center}
\begin{tabular}{ccc}
\includegraphics[width=0.38\textwidth]{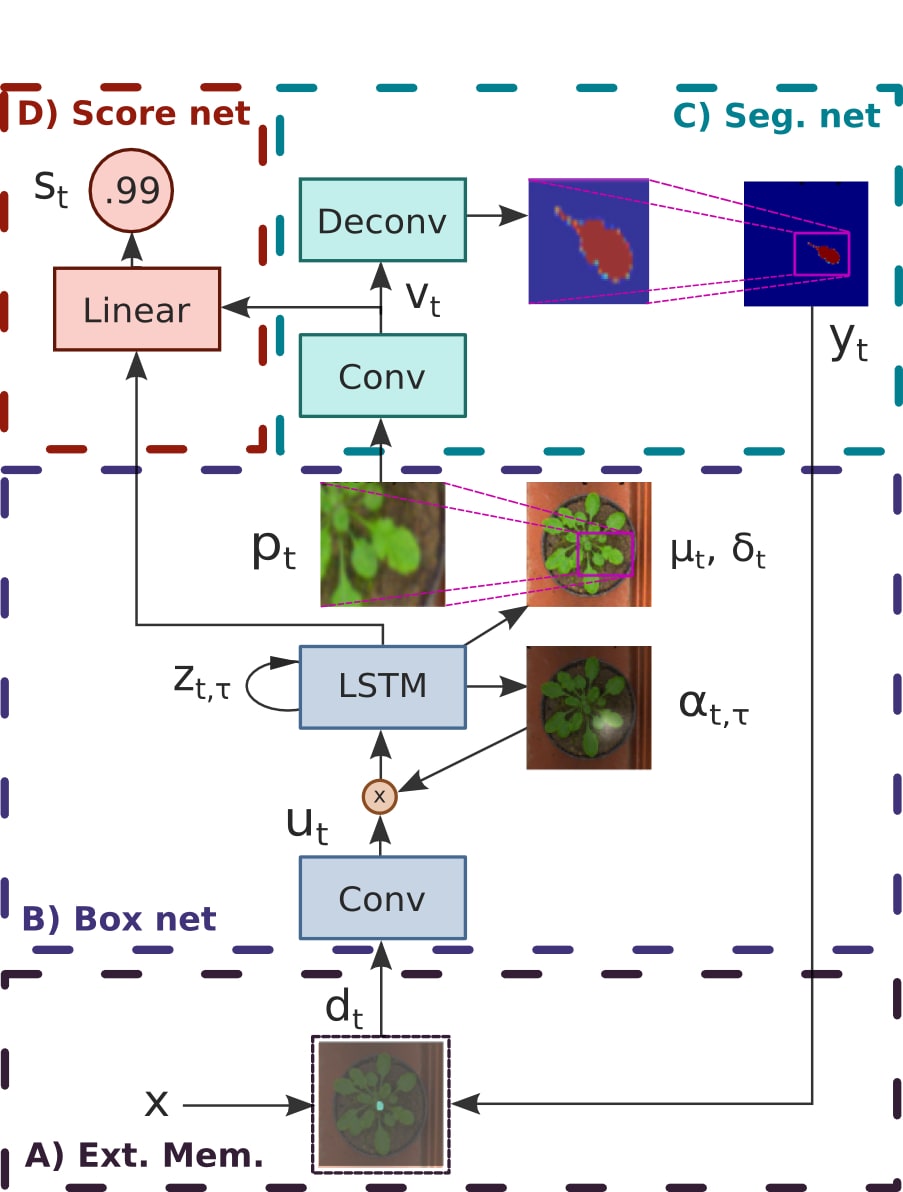} & \quad\quad\quad &
\includegraphics[width=0.27\textwidth]{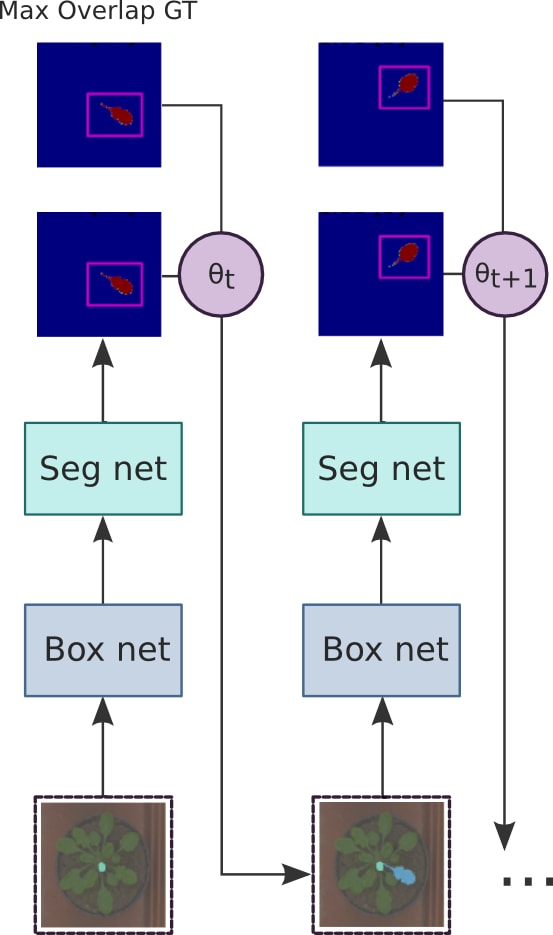}
\end{tabular}
\end{center}
\caption{
Left: Detailed network design.
Right: Sketch of training, and scheduled sampling;
during training, the weighting of
ground-truth instance segmentations
relative to model predictions
($\theta_t$) decays to zero.
}
\label{fig:model_arch}
\end{figure*}

Our proposed model has four major components:
A) an \*{external memory} that tracks the state of the segmented objects;
B) a \*{box proposal network} responsible for localizing objects of interest;
C) a \*{segmentation network}  for segmenting image pixels within the box; and
D) a \*{scoring network} that determines if an object instance has been found,
and also decides when to stop. See Figure~\ref{fig:model_arch} for an
illustration of these components.

\textbf{Notation}. We use the following notation to describe the model
architecture:
$\*{x}_0 \in \mathbb{R}^{H \times W \times C}$
is the input image
($H$, $W$ denotes the dimension, and $C$ denotes the color channel);
$t$ indexes the iterations of the model, and 
$\tau$ indexes the glimpses of the box network's inner RNN;
$\*y   = \{\*{y}_t | \*{y}_t \in [0, 1]^{H \times W}\}_{t=1}^T$,
$\*y^* = \{\*{y}^*_t | \*{y}^*_t \in \{0, 1\}^{H \times W}\}_{t=1}^T$
are the output/ground-truth segmentation sequences;
$\*s   = \{s_{t} | s_t \in [0, 1]\}_{t=1}^{T}$, 
$\*s^* = \{s^*_{t} | s^*_{t} \in \{0, 1\}\}_{t=1}^{T}$ 
are the output/ground-truth confidence score sequences.
$\*h = \text{CNN}(I)$
denotes passing an image $I$ through a CNN and returning the hidden activation
$\*h$. 
$I' = \text{D-CNN}(\*h)$
denotes passing an activation map $h$ through a de-convolutional network
(D-CNN) and returning an image $I'$.
$\*h_t = \text{LSTM}(\*h_{t-1}, \*x_t)$ 
denotes unrolling the long short-term memory (LSTM) 
by one time-step with the previous hidden state $\*h_{t-1}$ and
current input $\*x_t$, and returning the current hidden state $\*h_t$.
$\*h = \text{MLP}(\*x)$
denotes passing an input $\*x$ through a multi-layer perceptron (MLP) and
returning the hidden state $\*h$.

\begin{figure*}
\begin{center}
\resizebox{0.9\textwidth}{!}{
\begin{tabular}{ccc}
\includegraphics[width=0.3\textwidth]{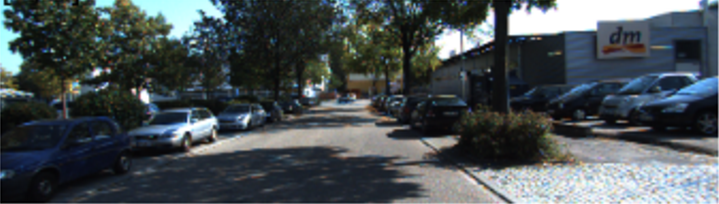} &
\includegraphics[width=0.3\textwidth]{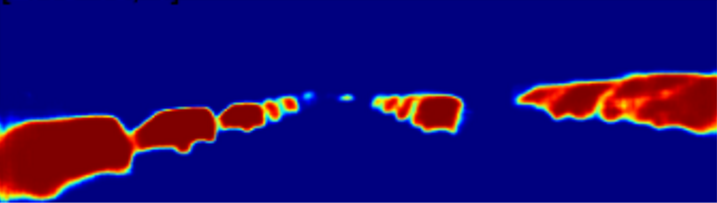} &
\includegraphics[width=0.3\textwidth]{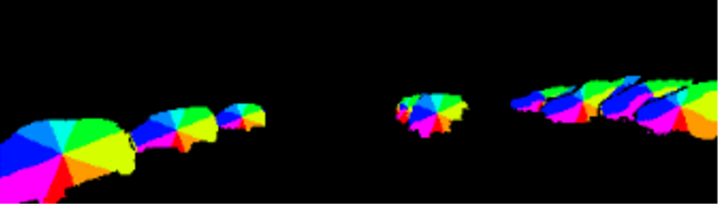}
\end{tabular}
}
\caption{Illustration of the output of the pretrained FCN. Left: input image.
Middle: predicted foreground. Right: predicted angle map.}
\label{fig:fcn}
\end{center}
\end{figure*}

\textbf{Input pre-processing}. We pre-train a FCN \cite{long15fcn} to perform
input pre-processing. This pre-trained FCN has two output components. The first is
a 1-channel pixel-level foreground segmentation, produced by a variant of the
DeconvNet \cite{noh15deconv} with skip connections. In addition to predicting
this foreground mask, as a second component we followed the work of Uhrig et al.
\cite{uhrig16insseg} by producing an angle map for each object. For each
foreground pixel, we calculate its relative angle towards the centroid of the
object, and quantize the angle into 8 different classes, forming 8 channels, as
shown in Figure~\ref{fig:fcn}. Predicting the angle map forces the model to
encode more detailed information about object boundaries. The architecture and
training of these components are detailed in the Appendix.
We denote $\*{x}_0$ as the original image (3 channel RGB), and $\*{x}$ as the
pre-processed image (9 channels: 1 for foreground and 8 for angles).

%%%%%%%%%%%%%%%%%%%%%%%%%%%%%%%%%%%%%%%%%%%%%%%%%%%%%%%%%%%%%%%%%%%%%%%%%%%%%%%
%%%%%%%%%%%%%%%%%%%%%%%%%%%%%%%%%%%%%%%%%%%%%%%%%%%%%%%%%%%%%%%%%%%%%%%%%%%%%%%
\subsection{Part A: External memory}

To decide where to look next based on the already segmented objects, we
incorporate an \*{external memory}, which provides object boundary details from
all previous steps. 
We hypothesize that providing
information of the completed segmentation helps the network reason about
occluded objects and determine the next region of interest. The canvas has 10
channels in total: the first channel of the canvas keeps adding new pixels from
the output of the previous time step, and the other channels store the 
input image.
\begin{align}
&\*{c}_t                 = \begin{cases}
                                \*{0},              & \text{if } t = 0       \\
                                \max(\*{c}_{t-1}, 
                                \*{y}_{t-1}),       & \text{otherwise}
                           \end{cases}                                       \\
&\*{d}_t = \left[\*{c}_{t}, \*{x}\right]
\end{align}

%%%%%%%%%%%%%%%%%%%%%%%%%%%%%%%%%%%%%%%%%%%%%%%%%%%%%%%%%%%%%%%%%%%%%%%%%%%%%%%
%%%%%%%%%%%%%%%%%%%%%%%%%%%%%%%%%%%%%%%%%%%%%%%%%%%%%%%%%%%%%%%%%%%%%%%%%%%%%%%
\subsection{Part B: Box network}

The box network plays a critical role, localizing the next object of interest.
The CNN in the box network outputs a $H' \times W' \times L$ feature map
$\*{u}_{t}$ ($H'$ is the height; $W'$ is the width; $L$ is the feature
dimension). CNN activation based on the entire image is too complex and
inefficient to process simultaneously. Simple pooling does not preserve
location; instead we employ a ``soft-attention'' (dynamic pooling) mechanism
here to extract useful information along spatial dimensions, weighted by
$\alpha_t^{h, w}$. Since a single glimpse may not give the upper network enough
information to decide where exactly to draw the box, we allow the glimpse LSTM
to look at different locations by feeding a dimension $L$ vector each time.
$\alpha$ is initialized to be uniform over all locations, and $\tau$ indexes the
glimpses.
\begin{align}
\*u_t            &= \text{CNN}(\*d_t)                                        \\
\*z_{t, \tau}    &= \begin{cases}
                    \*0,                         & \text{if } \tau = 0       \\
                    \text{LSTM}(\*z_{t, \tau-1},
                    \sum\limits_{h, w} \alpha^{h, w}_{t, \tau - 1} 
                    u^{h,w,l}_{t})               & \text{otherwise}
                    \end{cases}\\
\alpha_{t, \tau}^{h, w} &= \begin{cases}
                        1/(H' \times W'),        & \text{if } \tau = 0       \\
                        \text{MLP}(\*z_{t, \tau}), & \text{otherwise}
                        \end{cases}
\end{align}

We pass the LSTM's hidden state through a linear layer to obtain predicted box
coordinates. We parameterize the box by its normalized center
$(\tilde{g}_X,\tilde{g}_Y)$, and size $(\log \tilde{\delta}_X, \log
\tilde{\delta}_Y)$. A scaling factor $\gamma$ is also predicted by the linear
layer, and used when re-projecting the patch to the original image size.
\begin{align}
[\tilde{g}_{X,Y},
 \log \tilde{\delta}_{X,Y},
 \log \sigma_{X,Y},
 \gamma]                  &=  \*{w}_b^\top \*{z}_{t, \text{end}} + w_{b0}
\end{align}
\vspace{-0.4in}
\begin{align}
g_X      &= (\tilde{g}_X+1)W/2                                               \\
g_Y      &= (\tilde{g}_Y+1)H/2                                               \\
\delta_X &= \tilde{\delta}_X W                                               \\
\delta_Y &= \tilde{\delta}_Y H
\end{align}

\textbf{Extracting a sub-region}. We follow DRAW \cite{gregor15draw} and use a
Gaussian interpolation kernel to extract an $\tilde{H} \times \tilde{W}$ patch
from the $\tilde{x}$, a concatenation of the original image with $\*d_t$. 
We further allow the model to output rectangular patches to account for different
shapes of the object. $i, j$ index the location in the patch of dimension
$\tilde{H} \times \tilde{W}$, and $a, b$ index the location in the original
image. $F_X$ and $F_Y$ are matrices of dimension $W \times \tilde{W}$ and $H
\times \tilde{H}$, which indicates the contribution of the location $(a,b)$ in
the original image towards the location $(i,j)$ in the extracted patch.
$\mu_{X,Y}$ and $\sigma_{X,Y}$ are mean and variance of the Gaussian
interpolation kernel, predicted by the box network.
\begin{align}
\mu_X^i         &= g_X + (\delta_X + 1) \cdot
                   (i - \tilde{W} / 2 + 0.5) / \tilde{W}                     \\
\mu_Y^j         &= g_Y + (\delta_Y + 1) \cdot
                   (j - \tilde{H} / 2 + 0.5) / \tilde{H}                     \\
F_X^{a, i}      &= \frac{1}{\sqrt{2\pi} \sigma_X}                             
                   \exp \left(- \frac{(a - \mu_X^i)^2}{2\sigma_X^2} \right)  \\
F_Y^{b, j}      &= \frac{1}{\sqrt{2\pi} \sigma_Y}
                   \exp \left(- \frac{(b - \mu_Y^j)^2}{2\sigma_Y^2} \right)
\end{align}
\begin{align}
% \tilde{\*x}_t   &= [\*{x}_0, \*{d}_t^{\text{canvas}}]                      \\
\tilde{\*x}_t   &= [\*{x}_0, \*{d}_t]                        \\
\*p_t           &= \text{Extract}(\tilde{\*{x}}_t, F_Y, F_X) 
            \equiv F_Y^\top \tilde{\*{x}}_t F_X
\end{align}

%%%%%%%%%%%%%%%%%%%%%%%%%%%%%%%%%%%%%%%%%%%%%%%%%%%%%%%%%%%%%%%%%%%%%%%%%%%%%%%
%%%%%%%%%%%%%%%%%%%%%%%%%%%%%%%%%%%%%%%%%%%%%%%%%%%%%%%%%%%%%%%%%%%%%%%%%%%%%%%
\subsection{Part C: Segmentation network}

The remaining task is to segment out the pixels that belong to the dominant
object within the window. The segmentation network first utilizes a convolution
network to produce a feature map $\*v_t$. We then adopt a variant of the
DeconvNet \cite{noh15deconv} with skip connections, which appends deconvolution
(or convolution transpose) layers to upsample the low-resolution feature map to
a full-size segmentation. After the fully convolutional layers we get a
patch-level segmentation prediction heat map $\tilde{y}_t$. We then re-project
this patch prediction to the original image using the transpose of the
previously computed Gaussian filters; the learned $\gamma$ magnifies the signal
within the bounding box, and a constant $\beta$ suppresses the pixels outside
the box. Applying the sigmoid function produces segmentation values
between 0 and 1.
\begin{align}
\*v_t         &= \text{CNN}(\*p_t)                                         \\
\tilde{\*y}_t &= \text{D-CNN}(\*v_t)                                       \\
\*y_t         &= \text{sigmoid} \left(\gamma \cdot 
                 \text{Extract}(\tilde{\*y}_t, F_Y^\top, F_X^\top)
                 -\beta \right)
\end{align}

%%%%%%%%%%%%%%%%%%%%%%%%%%%%%%%%%%%%%%%%%%%%%%%%%%%%%%%%%%%%%%%%%%%%%%%%%%%%%%%
%%%%%%%%%%%%%%%%%%%%%%%%%%%%%%%%%%%%%%%%%%%%%%%%%%%%%%%%%%%%%%%%%%%%%%%%%%%%%%%
\subsection{Part D: Scoring network}

To estimate the number of objects in the image, and to terminate our sequential
process, we incorporate a scoring network, similar to the one presented in
\cite{romeraparedes15ris}. Our scoring network takes information from the
hidden states of both the box network ($\*z_t$) and segmentation network ($\*v_t$)
to produce a score between 0 and 1.
\begin{equation}
s_{t} = \text{sigmoid}(\*w_{zs}^\top\*z_{t, \text{end}} + 
                       \*w_{vs}^\top\*v_t + w_{s0})
\end{equation}

\textbf{Termination condition}. We train the entire model with a sequence
length determined by the maximum number of objects plus one. During inference,
we cut off iterations once the output score goes below 0.5. The score loss function
(described below) encourages scores to decrease monotonically.

%%%%%%%%%%%%%%%%%%%%%%%%%%%%%%%%%%%%%%%%%%%%%%%%%%%%%%%%%%%%%%%%%%%%%%%%%%%%%%%
%%%%%%%%%%%%%%%%%%%%%%%%%%%%%%%%%%%%%%%%%%%%%%%%%%%%%%%%%%%%%%%%%%%%%%%%%%%%%%%
\subsection{Loss functions}

\textbf{Joint loss}. The total loss function is a sum of three losses: the
segmentation matching IoU loss $\mathcal{L}_y$; the box IoU loss
$\mathcal{L}_b$; and the score cross-entropy loss $\mathcal{L}_s$:
\begin{equation}
\mathcal{L}(\*y, \*b, \*s) = \mathcal{L}_y(\*y, \*y^*) + 
                             \lambda_b \mathcal{L}_b(\*b, \*b^*) + 
                             \lambda_s \mathcal{L}_s(\*s, \*s^*)
\end{equation}
We fix the loss coefficients $\lambda_b$ and $\lambda_s$ to be 1 for all of our
experiments.

\textbf{(a) Matching IoU loss (mIOU)}. A primary challenge of instance
segmentation involves matching model and ground-truth instances.  We compute a
maximum-weighted  bipartite graph matching between the output instances and
ground-truth instances (c.f., \cite{stewart15lstmdet} and
\cite{romeraparedes15ris}). Matching makes the loss insensitive to the ordering
of the ground-truth instances. Unlike coverage scores proposed in
\cite{silberman14insseg} it directly penalizes both false positive and false
negative segmentations. The matching weight $\mathcal{M}_{i, j}$ is the IoU
score between a pair of segmentations. We use the Hungarian algorithm to compute
the matching; we do not back-propagate the network gradients through this
algorithm.
\begin{align}
\mathcal{M}_{i, j}        &= \text{softIOU}(\*y_i, \*y^*_j) 
                      \equiv \frac{\sum \*y_i \cdot \*y^*_j}
                             {\sum \*y_i + \*y^*_j - \*y_i \cdot \*y^*_j}    \\
\mathcal{L}_y(\*y, \*y^*) &= -\text{mIOU}(\*y, \*y^*)                        \\
                     &\equiv -\frac{1}{N} \sum_{i, j} 
                             \mathcal{M}_{i, j}
                             \mathbbm{1}[\text{match}(\*y_i) = \*y^*_j]
\end{align}

\textbf{(b) Soft box IoU loss}. Although the exact IoU can be derived from the
4-d box coordinates, its gradient vanishes when two boxes do not overlap, which
can be problematic for gradient-based  learning. Instead, we propose a soft
version of the box IoU. We use the same Gaussian filter to re-project a
constant patch on the original image, pad the ground-truth boxes, and then 
compute the mIOU between the predicted box and the matched padded ground-truth 
bounding box that is  scaled proportionally in both height and width.
\begin{align}
\*b_{t}                   &= \text{sigmoid}(\gamma \cdot
                             \text{Extract}(
                             \mathbf{1}, F_Y^\top, F_X^\top) - \beta)
\end{align}
\vspace{-0.3in}
\begin{align}
\mathcal{L}_b(\*b, \*b^*) &= -\text{mIOU}(\*b, \text{Pad}(\*b^*))
\end{align}

\textbf{(c) Monotonic score loss}. To facilitate automatic termination, the
network should output more confident objects first. We formulate a loss function
that encourages monotonically decreasing values in the score output. Iterations
with target score 1 are compared to the lower bound of preceding scores, and 0
targets to the upper bound of subsequent scores.
\begin{equation}
\begin{split}
\mathcal{L}_s(\*s, \*s^*) = \frac{1}{T}\sum_t 
          & -s^*_t \log\left(\min_{t' \le t} \{s_{t'}\}\right)    \\
          &-(1 - s^*_t) \log\left(1 - \max_{t' \ge t} \{s_{t'}\}\right)
\end{split}
\end{equation}

%%%%%%%%%%%%%%%%%%%%%%%%%%%%%%%%%%%%%%%%%%%%%%%%%%%%%%%%%%%%%%%%%%%%%%%%%%%%%%%
%%%%%%%%%%%%%%%%%%%%%%%%%%%%%%%%%%%%%%%%%%%%%%%%%%%%%%%%%%%%%%%%%%%%%%%%%%%%%%%
\subsection{Training procedure}
\label{sec:postproc}
\textbf{Bootstrap training}. The box and segmentation networks rely on the
output of each other to make decisions for the next time-step. Due to the
coupled nature of the two networks, we propose a bootstrap training procedure:
these networks are pre-trained with ground-truth segmentation and boxes, 
respectively, and in later stages we replace the ground-truth with the model 
predicted values.

\textbf{Scheduled sampling}. To smooth out the transition between stages, we
explore the idea of ``scheduled sampling'' \cite{bengio15schedsamp}, where we
gradually remove the reliance on ground-truth segmentation at the input of the
network. As shown in Figure~\ref{fig:model_arch}, during training there is a
stochastic switch ($\theta_t$) in the input of the external memory, to utilize
either the maximally overlapping ground-truth instance segmentation, or the
output of the network from the previous time step. By the end of the training,
the model completely relies on its own output from the previous step,  which
matches the test-time inference procedure.

\section{Related Work}

Instance segmentation has recently received a burst of research attention, as
it provides higher level of precision of image understanding compared to object
detection and semantic segmentation.

\textbf{Detector-based approaches}. Early work on object segmentation
\cite{borenstein02topdown} starts from a trained class detectors, which provide
a bottom-up merging criterion based on top-down detections. \cite{yang12layer}
extends this approach with a DPM detector, and uses layered graphical models to
reason about instance separation and occlusion. Besides boxes, models can also
leverage region proposals and region descriptors \cite{carreira15region}.
Simultaneous detection and segmentation (SDS) methods
\cite{hariharan14simultaneous,hariharan15hypercolumn,liang15rrinsseg,
li16iterative} apply newer CNN-based detectors such as RCNN
\cite{girshick14rcnn}, and perform segmentation using the bounding boxes as
starting point. \cite{li16iterative} proposes a trainable iterative refining
procedure.  Dai et al.~\cite{dai15insaware} propose a pipeline-based approach
which first predicts bounding box proposals and then performs segmentation
within each ROI.  Similarly, we jointly learn a detector and a segmentor;
however, we introduce a direct feedback loop in the sequence of predictions. In
addition, we learn the sequence ordering.

\textbf{Graphical model approaches}. Another line of research is to use
generative graphical model to express the dependency structure among instances
and pixels. Eslami et al. \cite{eslami16shapebm} proposes a restricted
Boltzmann machine to capture high-order pixel interactions for shape modelling.
A multi- stage pipeline proposed by Silberman et al. \cite{silberman14insseg}
is composed of patch-wise features based on deep learning, combined into a
segmentation tree.  More recently, Zhang et al. \cite{zhang16insseg} formulate
a dense CRF for instance segmentation; they apply a CNN on dense image patches
to make local predictions, and construct a dense CRF to produce globally
consistent labellings. Their key contribution is a shifting-label potential
that encourages consistency across different patches.  The graphical model
formulation entails long running times, and their energy functions are
dependent on instances being connected and having a clear depth ordering.

\textbf{Fully convolutional approaches}.  Fully convolutional networks
\cite{long15fcn} has emerged as a powerful tool to directly predict dense pixel
labellings. Pinheiro et al.  ~\cite{pinheiro15seg,pinheiro16refine} train a CNN
to generate object segmentation proposals, which runs densely on all windows at
multiple scales, a nd Dai et al.~\cite{dai16insfcn} uses relative location as
additional pixel labels. The output of both systems are object proposals, which
require further processing to get final segmentations. Other approaches using
FCNs are proposal-free, but rely on a bottom-up merging process.  Liang et
al.~\cite{liang15pfinsseg} predict dense pixel prediction of object location
and size, using clustering as a post-processing step. Uhrig et al.
~\cite{uhrig16insseg} present another approach based on FCNs, which is trained
to produces a semantic segmentation as well as an instance-aware angle map,
encoding the instance centroids.  Post-processing based on template matching
and instance fusion produces the instance identities. Importantly, they also
used ground-truth depth labels in training.  Concurrent work
\cite{bai16dwt,hayder16sais,kirillov16inscut} also explores a similar idea of
using FCNs to output instance-sensitive embeddings.

\textbf{RNN approaches}. Another recent line of research, e.g.,
\cite{stewart15lstmdet, park15ris, romeraparedes15ris} employs end-to-end
recurrent neural networks (RNN) to perform object detection and segmentation.
The sequential decomposition idea for structured prediction is also explored in
\cite{daume09searn, banica15seg}. A permutation agnostic loss function based on
maximum weighted bipartite matching was proposed by \cite{stewart15lstmdet}. To
process an entire image, they treat each element of a CNN feature map
individually. Similarly, our box proposal network also uses an RNN to generate
box proposals: instead of running hundreds of RNN iterations, we only run it
for a small number of iterations using a soft attention mechanism
\cite{xu15caption}. Romera-Paredes and Torr \cite{romeraparedes15ris} use
convolutional LSTM (ConvLSTM) \cite{shi2015convlstm} to produce instance
segmentation directly. However, since their ConvLSTM is required to handle
object detection, inhibition, and segmentation all convolutionally on a global
scale, and it is hard for their model to inhibit far apart instances. In
contrast, our architecture incorporates direct feedback from the prediction of
the previous instance, providing precise boundary inhibition, and our box
network confines the instance-wise segmentation within a local window.

\section{Experiments}

We refer readers to the Appendix for hyper-parameters and other training details
of the experiments \footnote{Our code is released at:
\url{https://github.com/renmengye/rec-attend-public}}.

\subsection{Datasets \& Evaluation}

\textbf{CVPPP leaf segmentation}. One instance segmentation benchmark is the
CVPPP plant leaf dataset \cite{minervini14cvppp}, which was developed due to the
importance of instance segmentation in plant phenotyping. We ran the A1 subset
of CVPPP plant leaf segmentation dataset. We trained our model on 128 labeled
images, and report results on the 33 test images. We compare our performance to
\cite{romeraparedes15ris}, and other top approaches that were published with the
CVPPP conference; see the collation study \cite{scharr16leaf} for details of
these  other approaches.

\input{figs/cvppp_fig.tex.part}
\input{figs/coco_zebra_fig.tex.part}

\textbf{KITTI car segmentation}. Instance segmentation also provides rich
information in the context of autonomous driving. Following \cite{zhang15insseg,
zhang16insseg, uhrig16insseg}, we also evaluated the model performance on the
KITTI car segmentation dataset. We trained the model with 3,712 training images,
and report performance on 120 validation images and 144 test images.

\textbf{Cityscapes}. Cityscapes provides multi-class instance-level annotation
and is currently the most comprehensive benchmark for instance segmentation,
containing 2,975 training images, 500 validation images, and 1,525 test images,
with 8 semantic classes (person, rider, car, truck, bus, train, motorcycle,
bicycle). We train our instance segmentation network as a class-agnostic model
for all classes, and apply a semantic segmentation mask obtained from
\cite{ghiasi16lrr} on top of our instance output. Since ``car'' is the most
common class, we report both average score on all classes, and individual score
on the ``car'' class.

\textbf{MS-COCO}. To test the effectiveness and portability of our algorithm, we
train our model on a subset of MS-COCO. As an initial study we select images of
zebras, and train on 1000 images. Since there are no methods that are directly
comparable, we leave the quantatitive results for Appendix.

\textbf{Ablation studies}. We also examine the relative importance of model
components via ablation studies, and report validation performance on the KITTI
dataset.
\begin{itemize}
\item \textbf{No pre-processing}. This network is trained to take as input raw 
image pixels, without the foreground segmentation or the angle map.
\vspace{-0.1in}
\item \textbf{No box net}. Instead of predicting segmentation within a box, the
output dimension of the segmentation network is the full image.
\vspace{-0.1in}
\item \textbf{No angles}. The pre-processor predicts the foreground segmentation
only, without the angle map.
\vspace{-0.1in}
\item \textbf{No scheduled sampling}. This network has the same architecture 
but trained without scheduled sampling (see Section~\ref{sec:postproc}), i.e.,
at training time, always use the maximum overlapped ground-truth.
\vspace{-0.1in}
\item \textbf{Fewer iterations}. The box network has fewer glimpses on
the convnet feature map (fewer LSTM iterations).
\end{itemize}

\begin{table}
% \begin{minipage}[t]{.5\textwidth}
\begin{minipage}[t]{\columnwidth}
\caption{Leaf segmentation and counting performance, averaged over all test
images, with standard deviation in parentheses}
\label{tab:cvppp}
\centering
\begin{small}

% \resizebox{1.0\columnwidth}{!}{
\begin{tabular}{l|ll}
\toprule
\toprule
                                  & SBD \u         & $|\text{DiC}|$ \d \\
\midrule
RIS+CRF \cite{romeraparedes15ris} & 66.6 (8.7)     & 1.1 (0.9)         \\
MSU \cite{scharr16leaf}           & 66.7 (7.6)     & 2.3 (1.6)         \\
Nottingham \cite{scharr16leaf}    & 68.3 (6.3)     & 3.8 (2.0)         \\
Wageningen \cite{yin14leaf}       & 71.1 (6.2)     & 2.2 (1.6)         \\
IPK \cite{pape14ipk}              & 74.4 (4.3)     & 2.6 (1.8)         \\
PRIAn \cite{giuffrida15count}     & -              & 1.3(1.2)          \\
\midrule
% Ours (FCN + Canvas)               & 79.3 (9.2)     & 0.9 (1.0)       \\
% Ours (FCN + ConvLSTM)             & 80.4 (8.9)     & 1.0 (0.8)       \\
Ours                              & \*{84.9} (4.8) & \*{0.8} (1.0)     \\
\bottomrule
\end{tabular}
% }
\end{small}
\end{minipage}
% \qquad
\end{table}
\vspace{10pt}

% \begin{table*}[t]
% \begin{minipage}[t]{\textwidth}
% \caption{KITTI vehicle segmentation results}
% \label{tab:kitti}
% \centering
% \begin{small}
% % Resize box start
% % \resizebox{\textwidth}{!}{
% \begin{tabular}{l|lllll}
% \toprule
% \toprule
%                      & FG IoU \u & MWCov \u & MUCov \u & AvgFP \d  & AvgFN \d  \\
% \midrule    
% DenseCRFv1
% \cite{zhang15insseg} & 77.3      & 70.9     & 52.2     & 0.597     & 0.736     \\
% DenseCRFv2
% \cite{zhang16insseg} & 78.5      & 74.1     & 55.2     & 0.417     & 0.833     \\
% FCN+Depth
% \cite{uhrig16insseg} & 84.1      & 79.7     & \*{75.8} & 0.201     & \*{0.159} \\
% \midrule
% Ours                 & \*{87.4}  & \*{83.2} & 69.9     & \*{0.118} & 0.278     \\
% \bottomrule
% \end{tabular}
% % }
% % Resize box end
% \end{small}
% \end{minipage}
% \end{table*}
\begin{table}[t]
% \begin{minipage}[t]{\textwidth}
\caption{KITTI vehicle segmentation results}
\label{tab:kitti}
\centering
\begin{small}

\resizebox{\columnwidth}{!}{
\begin{tabular}{l|l|llll}
\toprule
\toprule
                     & Set  & MWCov    & MUCov    & AvgFP     & AvgFN     \\
\midrule                                                                          
DepthOrder                                                                          
\cite{zhang15insseg} & test & 70.9     & 52.2     & 0.597     & 0.736     \\
DenseCRF                                                                        
\cite{zhang16insseg} & test & 74.1     & 55.2     & 0.417     & 0.833     \\
AngleFCN+Depth                                                                      
\cite{uhrig16insseg} & test & 79.7     & \*{75.8} & \*{0.201} & \*{0.159} \\
\midrule                                                                      
Ours                 & test & \*{80.0} & 66.9     & 0.764     & 0.201     \\
% Ours                 & test & \*{80.0} & 67.2     & 0.771     & 0.229     \\
% Ours                 & test & \*{80.0} & 67.2     & 0.771     & 0.229     \\
% Ours                 & test & 79.7     & 67.6     & 0.854     & 0.208   \\
\bottomrule
\end{tabular}
}

% Resize box end
\end{small}
% \end{minipage}
\end{table}

\textbf{Evaluation metrics}. We evaluate based on the  metrics used by the other
studies in the respective benchmarks. See Appendix for detailed equations for
these metrics.

For segmentation, symmetric best dice (SBD) is used for leaves. 
% (see Supp. Material and \cite{scharr16leaf} for detailed equations).
(see Equations~\ref{eq:bd}, \ref{eq:sbd}).
\begin{align}
\label{eq:bd}
\text{DICE}(A, B)                &= \frac{2 |A \hat B|}{|A| + |B|}    \\
\text{BD}(\{A_i\}, B)            &= \max_{i} \text{DICE}(A_i, B)
\end{align}
\begin{equation}
\begin{split}
\label{eq:sbd}
\text{SBD}(y_i\}, \{y^*_j\})     = \min (
                                    & \frac{1}{N} \sum_j 
                                    \text{BD}(\{y_i\}, y^*_j),        \\
                                    & \frac{1}{M} \sum_i
                                    \text{BD}(\{y^*_j\}, y_i) )
\end{split}
\end{equation}
Mean (weighted) coverage (MWCov, MUCov) are used for KITTI car segmentation.
%(see Supp. Material, \cite{zhang16insseg}).
(see Equations~\ref{eq:mwcov}, \ref{eq:mucov}). 
The coverage scores measure the instance-wise IoU for each ground-truth 
instance averaged over the image; MWCov further weights the score by the size 
of the ground-truth instance segmentation (larger objects get larger weights).
\begin{align}
\label{eq:mucov}
\text{MUCov}(\{y_i\}, \{y_j^*\}) &= \sum_i \frac{1}{N} 
                                    \max_j \text{IoU}(y_i, y_j^*)     \\
\label{eq:mwcov}
\text{MWCov}(\{y_i\}, \{y_j^*\}) &= \sum_i w_{\text{cov},i} 
                                    \max_j \text{IoU}(y_i, y_j^*)     \\
               w_{\text{cov},i}  &= \frac{|y_i|}{\sum_i |y_i|}
\end{align}
The Cityscapes evaluation uses average precision (AP),
%(see \cite{cordts16cityscapes}),
(see Equation~\ref{eq:ap}),
which counts the precision between a pair of matched prediction and 
groundtruth, for a range of IoU threshold values.
\begin{equation}
\begin{split}
\label{eq:ap}
\text{AP}(\{y_i\}, \{y_j^*\}) = &\max_s \sum_\theta \sum_j
                                Pr(y_{s(i)}, y_j) \cdot \\
                                &\mathbbm{1}
                                [\text{IoU}(y_{s(i)}, y^*_j) \ge \theta],
\end{split}
\end{equation}
where $s$ is a function that permutes the predicted instance order, and 
$\theta$ is the threshold from 0.5 to 0.95. 
Other scores include AP$^\text{50\%}$ for a threshold of $0.5$, and
AP$^\text{50m, 100m}$ for a subset of instances within 50m and 100m.

Counting is measured in absolute difference in count ($|\text{DiC}|$) 
(i.e., mean absolute error),
(see Equation~\ref{eq:dic}), 
average false positive (\text{AvgFP}), and average
false negative (\text{AvgFN}). False positive is the number of predicted
instances that do not overlap with the ground-truth, and false negative is the
number of  ground-truth instances that do not overlap with the prediction.
\begin{align}
\label{eq:dic}
|\text{DiC}| = \frac{1}{N}\sum_i |count_i - count_i^*|
\end{align}

\input{tabs/cityscapes_table.tex.part}
\begin{table}[t]
\caption{Ablation results on KITTI validation}
\label{tab:kitti-ablation}
\centering
\begin{small}
\resizebox{\columnwidth}{!}{
\begin{tabular}{l|l|llll}
\toprule
\toprule
                & Set &  MWCov   & MUCov    & AvgFP     & AvgFN     \\
\midrule                                                         
No Pre Proc.    & val &  55.6    & 45.0     & \*{0.125} & 0.417     \\  
No Box Net      & val &  57.0    & 49.1     & 0.757     & 0.375     \\ 
No Angle        & val &  71.2    & 63.3     & 0.542     & 0.342     \\
No Sched. Samp. & val &  73.6    & 63.9     & 0.350     & 0.317     \\
\midrule                                                         
Iter-1          & val &  64.1    & 54.8     & 0.200     & 0.375     \\
Iter-3          & val &  71.3    & 63.4     & 0.417     & 0.308     \\
\midrule                                                           
Full (Iter-5)   & val & \*{75.1} & \*{64.6} & 0.375     & \*{0.283} \\
\bottomrule
\end{tabular}
}
\end{small}
\end{table}

\input{figs/cityscapes_fig.tex.part}

\subsection{Results \& Discussion}
Example results on the leaf segmentation task are shown in
Figure~\ref{fig:cvppp_out}. On this task, our best model outperforms the
previous state-of-the-art by a large margin in both segmentation and counting
(see Table~\ref{tab:cvppp}). In particular, our method has significant
improvement over a previous RNN-based instance segmentation method
\cite{romeraparedes15ris}. We found that our model with the FCN pre-processor
overfit on this task, and we thus utilized the simpler version without input
pre-processing. This is not surprising, as the dataset is very small, and
including the FCN significantly increases the input dimension and number of
parameters.

On the KITTI task, Figure~\ref{fig:cityscapes_out} row 4-5 shows that our model
can segment cars in a wide variety of poses. Our method out-performs
\cite{zhang15insseg,zhang16insseg}, but scores lower than \cite{uhrig16insseg}
(see Table~\ref{tab:kitti}). One possible explanation is their inclusion of
depth information during training, which may help the model disambiguate distant
object boundaries. Moreover, their bottom-up ``instance fusion'' method plays a
crucial role (omitting this leads to a steep performance drop); this likely
helps segment smaller objects, whereas our box network does not reliably detect
distant cars.

The displayed segmentations demonstrate that our top-down attentional inference
is crucial for a good instance recognition model. This allows disconnected
components belonging to objects such as a bicycle
(Figure~\ref{fig:cityscapes_out} row 2) to be recognized as a whole piece
whereas \cite{zhang16insseg} models the connectedness as their energy potential.
Our model also shows impressive results in heavily occluded scenes, when only
small proportion of the car is visible (Figure~\ref{fig:cityscapes_out} row 1),
and when the zebra in the back reveals two disjoint pieces of its body
(Figure~\ref{fig:coco_zebra_out} right). Another advantage is that our model
directly outputs the final segmentation, eliminating the need for post-
processing, which is often required by other methods (e.g.
\cite{liang15pfinsseg, uhrig16insseg}).

Our method also learns to rank the importance of instances through visual
attention. During training, the model first attended to objects in a spatial
ordering (e.g. left to right), and then gradually shifted to a more
sophisticated ordering based on confidence, with larger attentional jumps in
between timesteps. 

Some failure cases of our approach, e.g., omitting distant objects and under-
segmentation, may be explained by our downsampling factor; to manage training
time, we down-sample KITTI and Cityscapes by a factor around 4, whereas
\cite{uhrig16insseg} does not do any any downsampling. We also observe a lack of
higher-order reasoning, e.g., the inclusion of the ``third'' limb of a person in
the bottom of Figure~\ref{fig:cityscapes_out}. As future work, these problems
can be addressed by a combination of our method with bottom-up merging methods
(e.g. \cite{liang15pfinsseg,uhrig16insseg}) and higher order graphical models
(e.g. \cite{nguyen14highcrf,eslami16shapebm}).

Finally, our ablation studies (see Table~\ref{tab:kitti-ablation}) help
elucidate the contribution of some model components. The initial foreground
segmentation, and the box network both play crucial roles , as seen in the
coverage measures. Scheduled sampling results in slightly better performance,
by making training resemble testing, gradually forcing the model to carry out a
full sequence.  Larger numbers of glimpses of the LSTM (Iter-$n$) helps the
model significantly by allowing more information to be considered before
outputting the next region of interest. Finally, we note that KITTI has a
fairly small validation and test set, so these results are highly variable (see
last lines of Table \ref{tab:kitti-ablation} versus Table \ref{tab:kitti}).

\section{Conclusion}

In this work, we borrow intuition from human counting and formulate instance
segmentation as a recurrent attentive process. Our end-to-end recurrent
architecture demonstrates significant improvement compared to earlier
formulations using RNN on the same tasks, and shows state-of-the-art results on
challenging instance segmentation datasets. We address the classic object
occlusion problem with an external memory, and the attention mechanism permits
segmentation at a fine resolution.

Our attentional architecture significantly reduces the number of parameters, and
the performance is quite strong despite being trained with only 100 leaf images
and under 3,000 road scene images. Since our model is end-to-end trainable and
does not depend on prior knowledge of the object type (e.g. size,
connectedness), we expect our method performance to scale directly with the
number of labelled images, which is certain to increase as this task gains in
popularity and new datasets become available. As future work, we are currently
extending our model to tackle highly multi-class instance segmentation, such as
the MS-COCO dataset, and more structured understanding of everyday scenes.

\vspace{-1em}
{\small 
\paragraph{Acknowledgements}
Supported by Samsung and the Intelligence Advanced Research Projects
Activity (IARPA) via Department of Interior/Interior Business Center (DoI/IBC)
contract number D16PC00003. The U.S. Government is authorized to reproduce and
distribute reprints for Governmental purposes notwithstanding any copyright
annotation thereon. Disclaimer: The views and conclusions contained herein are
those of the authors and should not be interpreted as necessarily representing
the official policies or endorsements, either expressed or implied, of IARPA,
DoI/IBC, or the U.S. Government.}
\vspace{-1em}
\input{references.tex.part}
\appendix

\section{Training procedure specification} 

We used the Adam optimizer \cite{kingma14adam} with learning rate 0.001 and
batch size of 8. The learning rate is multiplied by 0.85 for every 5000 steps
of training.

\subsection{Scheduled sampling}
We denote $\theta_t$ as the probability of feeding in ground-truth segmentation
that has the greatest overlap with the previous prediction, as opposed to model
output. $\theta_t$ decays exponentially as training proceeds, and for
larger $t$, the decay occurs later:
\vspace{-3pt}
\begin{align}
\theta_t &= 
\min \left(\Gamma_t \exp \left(-\frac{epoch - S}{S_2} \right), 1 \right) \\
\Gamma_t &= 1 + \log(1 + Kt)
\end{align}
where $epoch$ is the training index, $S$, $S_2$, and $K$ are constants. In the
experiments reported here, these values are $10000$, $2885$, and $3$.

\section{More experimental results}
We include the segmentation and counting performance on the MS-COCO zebra 
images in Table~\ref{tab:coco-count}. In terms of counting, our model 
out-performs a baseline method that runs an object detector and then
non-maximal suppression, and a new associative-subitizing method
~\cite{chattopadhyay16count}.

\begin{table}
\caption{MS-COCO Zebra Results}
\label{tab:coco-count}
\centering
\begin{small}
\begin{tabular}{l|llll}
\toprule
\toprule
                             & MWCov \u & MUCov \u & $|$DiC$|$ \d & Acc. \u  \\
\midrule                                                                       
detect                                                                         
\cite{chattopadhyay16count}  & -        & -        & 2.56         & -        \\
aso-sub                                                                        
\cite{chattopadhyay16count}  & -        & -        & 1.03         & -        \\
\midrule                                                                       
Ours                         & 69.2     & 64.2     & \*{0.79}     & 0.57     \\
\bottomrule
\end{tabular}
\end{small}
\end{table}

\section{Model architecture}

\subsection{Foreground + Orientation FCN}

We resize the image to uniform size. For CVPPP and MS-COCO dataset, we adopt a
uniform size of $224\times224$, for KITTI, we adopt $128\times448$, and for
Cityscapes $256\times512$ (4x downsampling).
Table~\ref{tab:fcn_spec} lists the specification of all layers.

\begin{table*}[h!]
\caption{FCN specification}
\label{tab:fcn_spec}
\centering
\resizebox{\textwidth}{!}{
\begin{small}
\begin{tabular}{ccccccc}
Name       & Type   & Input                & Spec (size/stride)             & Size CVPPP/MS-COCO      & Size KITTI             & Size Cityscapes        \\
input      & input  & -                    & -                              & $224\times224\times3$   & $128\times448\times3$  & $256\times512\times3$  \\
conv1-1    & conv   & input                & $3\times3\times3\times32$      & $224\times224\times32$  & $128\times448\times64$ & $256\times512\times64$ \\
conv1-2    & conv   & conv1-1              & $3\times3\times32\times64$     & $224\times224\times64$  & $128\times448\times32$ & $256\times512\times32$ \\
pool1      & pool   & conv1-2              & max $2\times2$                 & $112\times112\times64$  & $64\times224\times64$  & $128\times256\times64$  \\
conv2-1    & conv   & pool1                & $3\times3\times64\times64$     & $112\times112\times64$  & $64\times224\times64$  & $128\times256\times64$  \\
conv2-2    & conv   & conv2-1              & $3\times3\times64\times96$     & $112\times112\times96$  & $64\times224\times96$  & $128\times256\times96$  \\
pool2      & pool   & conv2-2              & max $2\times2$                 & $56\times56\times96$    & $32\times112\times96$  & $64\times128\times96$  \\
conv3-1    & conv   & pool2                & $3\times3\times96\times96$     & $56\times56\times96$    & $32\times112\times96$  & $64\times128\times96$  \\
conv3-2    & conv   & conv3-1              & $3\times3\times96\times128$    & $56\times56\times128$   & $32\times112\times128$ & $64\times128\times128$ \\
pool3      & pool   & conv3-2              & max $2\times2$                 & $28\times28\times128$   & $16\times56\times128$  & $32\times64\times128$  \\
conv4-1    & conv   & pool3                & $3\times3\times128\times128$   & $28\times28\times128$   & $16\times56\times128$  & $32\times64\times128$  \\
conv4-2    & conv   & conv4-1              & $3\times3\times128\times128$   & $28\times28\times128$   & $16\times56\times128$  & $32\times64\times128$  \\
conv4-3    & conv   & conv4-2              & $3\times3\times128\times128$   & $28\times28\times128$   & $16\times56\times128$  & $32\times64\times128$  \\
conv4-4    & conv   & conv4-3              & $3\times3\times128\times128$   & $28\times28\times128$   & $16\times56\times128$  & $32\times64\times128$  \\
conv4-5    & conv   & conv4-4              & $3\times3\times128\times128$   & $28\times28\times128$   & $16\times56\times128$  & $32\times64\times128$  \\
conv4-6    & conv   & conv4-5              & $3\times3\times128\times128$   & $28\times28\times128$   & $16\times56\times128$  & $32\times64\times128$  \\
conv4-7    & conv   & conv4-6              & $3\times3\times128\times128$   & $28\times28\times128$   & $16\times56\times128$  & $32\times64\times128$  \\
conv4-8    & conv   & conv4-7              & $3\times3\times128\times256$   & $28\times28\times256$   & $16\times56\times256$  & $32\times64\times256$  \\
pool4      & pool   & conv4-8              & max $2\times2$                 & $14\times14\times256$   & $8\times28\times256$   & $16\times32\times256$   \\
conv5-1    & conv   & pool4                & $3\times3\times256\times256$   & $14\times14\times256$   & $8\times28\times256$   & $16\times32\times256$   \\
conv5-2    & conv   & conv5-1              & $3\times3\times256\times256$   & $14\times14\times256$   & $8\times28\times256$   & $16\times32\times256$   \\
conv5-3    & conv   & conv5-2              & $3\times3\times256\times256$   & $14\times14\times256$   & $8\times28\times256$   & $16\times32\times256$   \\
conv5-4    & conv   & conv5-3              & $3\times3\times256\times512$   & $14\times14\times512$   & $8\times28\times512$   & $16\times32\times512$   \\
pool5      & pool   & conv5-4              & max $2\times2$                 & $7\times7\times512$     & $4\times14\times512$   & $8\times16\times512$   \\
deconv6-1  & deconv & pool5                & $3\times3\times256\times512/2$ & $14\times14\times256$   & $8\times28\times256$   & $16\times32\times256$   \\
deconv6-2  & deconv & deconv6-1 + conv5-3  & $3\times3\times256\times512$   & $14\times14\times256$   & $8\times28\times256$   & $16\times32\times256$   \\
deconv7-1  & deconv & deconv6-2            & $3\times3\times128\times256/2$ & $28\times28\times128$   & $16\times56\times128$  & $32\times64\times128$  \\
deconv7-2  & deconv & deconv7-1 + conv4-7  & $3\times3\times128\times256$   & $28\times28\times128$   & $16\times56\times128$  & $32\times64\times128$  \\
deconv8-1  & deconv & deconv7-2            & $3\times3\times96\times128/2$  & $56\times56\times96$    & $32\times112\times96$  & $64\times128\times96$  \\
deconv8-2  & deconv & deconv8-1 + conv3-1  & $3\times3\times96\times192$    & $56\times56\times96$    & $32\times112\times96$  & $64\times128\times96$  \\
deconv9-1  & deconv & deconv8-2            & $3\times3\times64\times96/2$   & $112\times112\times64$  & $64\times224\times64$  & $128\times256\times64$  \\
deconv9-2  & deconv & deconv9-1            & $3\times3\times64\times64$     & $112\times112\times64$  & $64\times224\times64$  & $128\times256\times64$  \\
deconv10-1 & deconv & deconv9-2            & $3\times3\times32\times64/2$   & $224\times224\times32$  & $128\times448\times32$ & $256\times512\times32$ \\
deconv10-2 & deconv & deconv10-1           & $3\times3\times32\times32$     & $224\times224\times32$  & $128\times448\times32$ & $256\times512\times32$ \\
deconv10-3 & deconv & deconv10-2 + input   & $3\times3\times9\times35$      & $224\times224\times9$   & $128\times448\times9$  & $256\times512\times9$  \\
\end{tabular}
\end{small}
}
\end{table*}

\subsection{External memory}

\begin{table*}[t]
\caption{External memory specification}
\label{tab:ext_mem_spec}
\centering
\begin{small}
\begin{tabular}{ccccccc}
Name       & Filter spec     & Size CVPPP/MS-COCO     & Size KITTI            & Size Cityscapes       \\
ConvLSTM   & $3\times3$      & $224\times224\times9$  & $128\times448\times9$ & $256\times512\times9$ \\
\end{tabular}
\end{small}
\end{table*}

\subsection{Box network}

The box network takes in 9 channels of input directly from the output of the 
FCN. It goes through a CNN structure again and uses the attention vector 
predicted by the LSTM to perform dynamic pooling in the last layer. The CNN 
hyperparameters are listed in Table~\ref{tab:box_cnn_spec} and the LSTM and 
glimpse MLP hyperparameters are listed in Table~\ref{tab:box_lstm_spec}. The 
glimpse MLP takes input from the hidden state of the LSTM and outputs a vector 
of normalized weighting over all the box CNN feature map spatial grids.

\begin{table*}[t]
\caption{Box network CNN specification}
\label{tab:box_cnn_spec}
\centering
% \resizebox{\columnwidth}{!}{
\begin{small}
\begin{tabular}{ccccccc}
Name    & Type  & Input    & Spec (size/stride)          & Size CVPPP/MS-COCO      & Size KITTI              & Size Cityscapes             \\
input   & input & -        & -                           & $224\times224\times9$   & $128\times448\times9$   & $256\times512\times9$   \\
conv1-1 & conv  & input    & $3\times3\times9\times16$   & $224\times224\times16$  & $128\times448\times16$  & $256\times512\times16$  \\
pool1   & pool  & conv1-2  & max $2\times2$              & $112\times112\times16$  & $64\times224\times16$   & $128\times256\times16$  \\
conv1-2 & conv  & conv1-1  & $3\times3\times16\times16$  & $112\times112\times16$  & $64\times224\times16$   & $128\times256\times16$  \\
pool1   & pool  & conv1-2  & max $2\times2$              & $56\times56\times16$    & $32\times112\times16$   & $64\times128\times16$   \\
conv2-1 & conv  & pool1    & $3\times3\times16\times32$  & $56\times56\times32$    & $32\times112\times32$   & $64\times128\times32$   \\
conv2-2 & conv  & conv2-1  & $3\times3\times32\times32$  & $56\times56\times32$    & $32\times112\times32$   & $64\times128\times32$   \\
pool2   & pool  & conv2-2  & max $2\times2$              & $28\times28\times32$    & $16\times56\times32$    & $32\times64\times32$    \\
conv3-1 & conv  & pool2    & $3\times3\times32\times64$  & $28\times28\times64$    & $16\times56\times64$    & $32\times64\times64$    \\
conv3-2 & conv  & conv3-1  & $3\times3\times64\times64$  & $28\times28\times64$    & $16\times56\times64$    & $32\times64\times64$    \\
pool3   & pool  & conv3-2  & max $2\times2$              & $14\times14\times64$    & $8\times28\times64$     & $16\times32\times64$    \\
conv3-1 & conv  & pool2    & $3\times3\times64\times64$  & $14\times14\times64$    & $8\times28\times64$     & $16\times32\times64$    \\
conv3-2 & conv  & conv3-1  & $3\times3\times64\times64$  & $14\times14\times64$    & $8\times28\times64$     & $16\times32\times64$    \\
pool3   & pool  & conv3-2  & max $2\times2$              & $7\times7\times64$      & $4\times14\times64$     & $8\times16\times64$     \\
\end{tabular}
\end{small}
% }
\end{table*}

\begin{table*}[t]
\caption{Box network LSTM specification}
\label{tab:box_lstm_spec}
\centering
\begin{small}
\begin{tabular}{cccccc}
Name        & Size CVPPP/MS-COCO     & Size KITTI   & Size Cityscapes  \\
LSTM        & $256$                  & $256$        & $256$            \\
GlimpseMLP1 & $256$                  & $256$        & $256$            \\
GlimpseMLP2 & $7\times7$             & $4\times14$  & $8\times16$      \\
\end{tabular}
\end{small}
\end{table*}

\subsection{Segmentation network}

The segmentation networks takes in a patch of size $48\times48$ with multiple
channels. The first three channels are the original image R, G, B channels.
Then there are 8 channels of orientation angles, and then 1 channel of
foreground heat map, all predicted by FCN. Full details are listed in
Table~\ref{tab:seg_net_spec}. Constant $\beta$ is chosen to be 5.

\begin{table*}[t]
\caption{Segmentation network specification}
\label{tab:seg_net_spec}
\centering
% \resizebox{\columnwidth}{!}{
\begin{small}
\begin{tabular}{cccccc}
Name       & Type   & Input                & Spec (size/stride)             & Size                 \\
input      & input  & -                    & -                              & $48\times48\times13$ \\
conv1-1    & conv   & input                & $3\times3\times13\times16$     & $48\times48\times16$ \\
conv1-2    & conv   & conv1-1              & $3\times3\times16\times32$     & $48\times48\times32$ \\
pool1      & pool   & conv1-2              & max $2\times2$                 & $24\times24\times32$ \\
conv2-1    & conv   & pool1                & $3\times3\times32\times32$     & $24\times24\times32$ \\
conv2-2    & conv   & conv2-1              & $3\times3\times32\times64$     & $24\times24\times64$ \\
pool3      & pool   & conv2-2              & max $2\times2$                 & $12\times12\times64$ \\
conv3-1    & conv   & pool2                & $3\times3\times64\times64$     & $12\times12\times64$ \\
conv3-2    & conv   & conv3-1              & $3\times3\times64\times96$     & $12\times12\times96$ \\
pool3      & pool   & conv3-2              & max $2\times2$                 & $6\times6\times96$   \\
deconv4-1  & deconv & pool3                & $3\times3\times64\times96/2$   & $12\times12\times64$ \\
deconv4-2  & deconv & deconv4-1 + conv3-1  & $3\times3\times64\times128$    & $12\times12\times64$ \\
deconv5-1  & deconv & deconv4-2 + conv2-2  & $3\times3\times32\times128/2$  & $24\times24\times32$ \\
deconv5-2  & deconv & deconv5-1 + conv2-1  & $3\times3\times32\times64$     & $24\times24\times32$ \\
deconv6-1  & deconv & deconv5-2 + conv1-2  & $3\times3\times16\times64/2$   & $48\times48\times16$ \\
deconv6-2  & deconv & deconv6-1 + conv1-1  & $3\times3\times16\times32$     & $48\times48\times16$ \\
deconv6-3  & deconv & deconv6-2 + input    & $3\times3\times1\times29$      & $48\times48\times1$  \\
\end{tabular}
\end{small}
% }
\end{table*}

\end{document}